# Robot Metabolism: Towards machines that can grow by consuming other machines

## Authors

Philippe Martin Wyder[1,2*], Riyaan Bakhda[3], Meiqi Zhao[3], Quinn A. Booth[4], Matthew E. Modi[4], Andrew Song[1], Simon Kang[1], Jiahao Wu[1], Priya Patel[1], Robert T. Kasumi[1], David Yi[1], Nihar Niraj Garg[1], Pranav Jhunjhunwala[1], Siddharth Bhutoria[3], Evan H. Tong[3], Yuhang Hu[1], Judah Goldfeder[3], Omer Mustel[3], Donghan Kim[3] and Hod Lipson[1]

## Affiliations

[1]Mechanical Engineering, Columbia University, 220 S. W. Mudd Building, 500 West 120th Street, New York, 10027, NY, USA.

[2]Applied Mathematics, University of Washington, Lewis Hall #201, Seattle, 98195, WA, USA.

[3]Computer Science, Columbia University, 450 S. W. Mudd Building, 500 West 120th Street, New York, 10027, NY, USA.

[4]Electrical Engineering, Columbia University, 1310G S. W. Mudd Building, 500 West 120th Street, New York, 10027, NY, USA.

*Corresponding author
E-mail: philippe.wyder@columbia.edu

## Abstract

Biological lifeforms can heal, grow, adapt, and reproduce -- abilities essential for sustained survival and development. In contrast, robots today are primarily monolithic machines with limited ability to self-repair, physically develop, or incorporate material from their environments. While robot minds rapidly evolve new behaviors through AI, their bodies remain closed systems, unable to systematically integrate material to grow or heal. We argue that open-ended physical adaptation is only possible when robots are designed using a small repertoire of simple modules. This allows machines to mechanically adapt by consuming parts from other machines or their surroundings and shed broken components. We demonstrate this principle on a truss modular robot platform. We show how robots can grow bigger, faster, and more capable by consuming materials from their environment and other robots. We suggest that machine metabolic processes like those demonstrated here will be an essential part of any sustained future robot ecology.

# Teaser

Robots grow bigger, faster and more capable by absorbing and integrating material in their environment.

# Introduction

Biological organisms operate as open systems: they absorb material from their environment and expel waste (*1*). This process is the basis for the long-term resilience of biological organisms over their lifetime(*2*, *3*). Progress in artificial intelligence has advanced robots' ability to adapt by learning new behaviors, but has left the robots' physical morphology fixed and monolithic. Typical robots today cannot increase in size and complexity, adapt, or self-repair. In contrast, biological lifeforms developed the ability for physical adaptation, repair, and replication, including absorbing and expelling material, long before any form of intelligence ever emerged(*4–6*). In light of that, artificial intelligence, although important, may just be one piece of the puzzle of true robot autonomy: robot self-sufficiency. For robots to become resilient and sustainable in the long term, we must develop processes that allow them to act as open systems and develop physically by consuming, expelling, and reusing material from their environment. We call this process *robot metabolism*.

Unlike traditional robot manufacturing processes, where robots may be involved in the process of making robots in a variety of ways, a robotic adaptation process qualifies as *robot metabolism* if it satisfies two criteria: First, *robot metabolism* cannot rely on active physical support from any external system to accomplish its growth; the robot must grow using only its own abilities. The only external assistance allowed is that which comes from other robots made of the same components. Second, the only external provision to *robot metabolism* is energy and material in the form of robots or robot parts. No new types of external components can be provided. In the case of the platform used in this work, material comes in the form of robot modules and energy in the form of electricity stored in each module's batteries.

The concept of *robot metabolism* raises more questions than we can answer in this paper. Thus, we focused on a set of key challenges: self-assembly, self-improvement, recombination after separation, and robot-to-robot assisted-reconfiguration. In this work, we demonstrate the potential of this approach and introduce a robot platform capable of achieving it. We believe that this is the first demonstration of a robot system that can grow from single parts into a full 3D robot, while systematically improving its own capability in the process and without requiring external machinery.

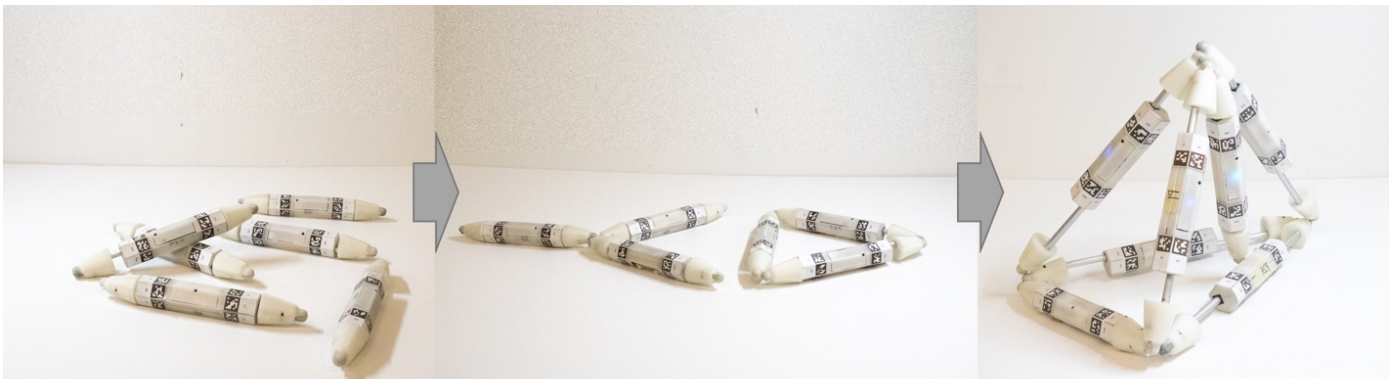

**Figure 1: Robot metabolism concept.** Robot modules can grow by consuming and reusing parts from their environment and other robots. This ability, essential to biological lifeforms, is crucial for developing a self-sustaining robot ecology. This paper demonstrates the above developmental sequence in detail: from individual modules to a fully assembled ratchet tetrahedron robot.

The choice of robotic building blocks is key as it spans the ultimate space for all possible designs. Biological lifeforms comprise only about 20 amino acids assembled into polypeptides during protein synthesis, ultimately giving rise to innumerable proteins and millions of self-sustaining lifeforms(*7*). Similarly, modular robots constructed from a finite set of simple, standardized components give rise to diverse functional structures and adaptive mechanisms. We believe that imitating nature's methods, rather than merely its results will lead more fundamental innovation in robotics. Replicating animals and humans in the form of robot dogs and humanoid robots is ultimately limiting. Thus, the robot building blocks need to be designed with the capacity for *robot metabolism.* Once developed, platforms capable of *robot metabolism* provide a physical counterpart to self-improving artificial intelligence. Thus, we open the possibility of robots changing their own form to ultimately overcome the limitations of human ingenuity.

We introduce the Truss Link, a robot building block designed to enable *robot metabolism*. The Truss Link is a simple, expandable, and contractible, bar-shaped robot module with two free-form magnetic connectors on each end. Animating any structure, Truss Links form robotic "organisms" that can grow by integrating material from their environment or from other robots (see Fig. 1). We show how two substructures can combine to form a larger robot, how two-dimensional (2D) structures can fold into three-dimensional (3D) shapes, how robot parts can be shed and then be replaced by another found part, and how one robot can help another "grow" through assisted reconfiguration. This work includes some results previously shared at the IEEE ReMar 2024 conference (*8*). In our conference paper we shared a limited, hardware-focused treatment of the Robot Link, i.e., the Truss Link without its free-form attachment/detachment mechanism. In this work, we present our Truss Link capable of Robot Metabolism, including its orientation agnostic, passively actuated, permanent magnet attachment/detachment mechanism, our simulation and corresponding quantitative results, the ratchet tetrahedron formation as the final stage of our developmental transformations, the improvement at every stage across all developmental transformations from Figure 3 in Movie S2, the shedding of dead links, and assisted tetrahedron formation.

**Truss Link**

Truss Links can be used to build modular robots. Modular robot systems comprise multiple parts called modules, links, or cells that can self-assemble or be assembled to achieve an objective. The Truss Link is the basic building block of our modular robot system. Toshio Fukada sparked a new generation of research, when he introduced modular robotics in 1988 (*9*). Modular robots promise increased versatility, configurability, scalability, resiliency and ability to self-reconfigure and evolve(*10–12*). Additionally, robot modularity could make robots cheaper if the modules were mass-produced(*10*). Modular robots are potentially resilient as a result of their redundancy and modularity, rather than mere material strength.

Modular robots can be classified as self-reconfiguring or manually reconfigurable robots(*13*). Self-reconfiguring robots can attach and detach from other modules automatically, while manually reconfigurable robots must be assembled by an operator. Truss Links enable modular self-configuring robots. A single Truss Link is capable only of motion in one dimension and, therefore, is limited to crawling forwards and backwards. Once a multi-link topology such as a triangle or tetrahedron has been formed, the system becomes fully controllable in 2D or 3D, respectively.

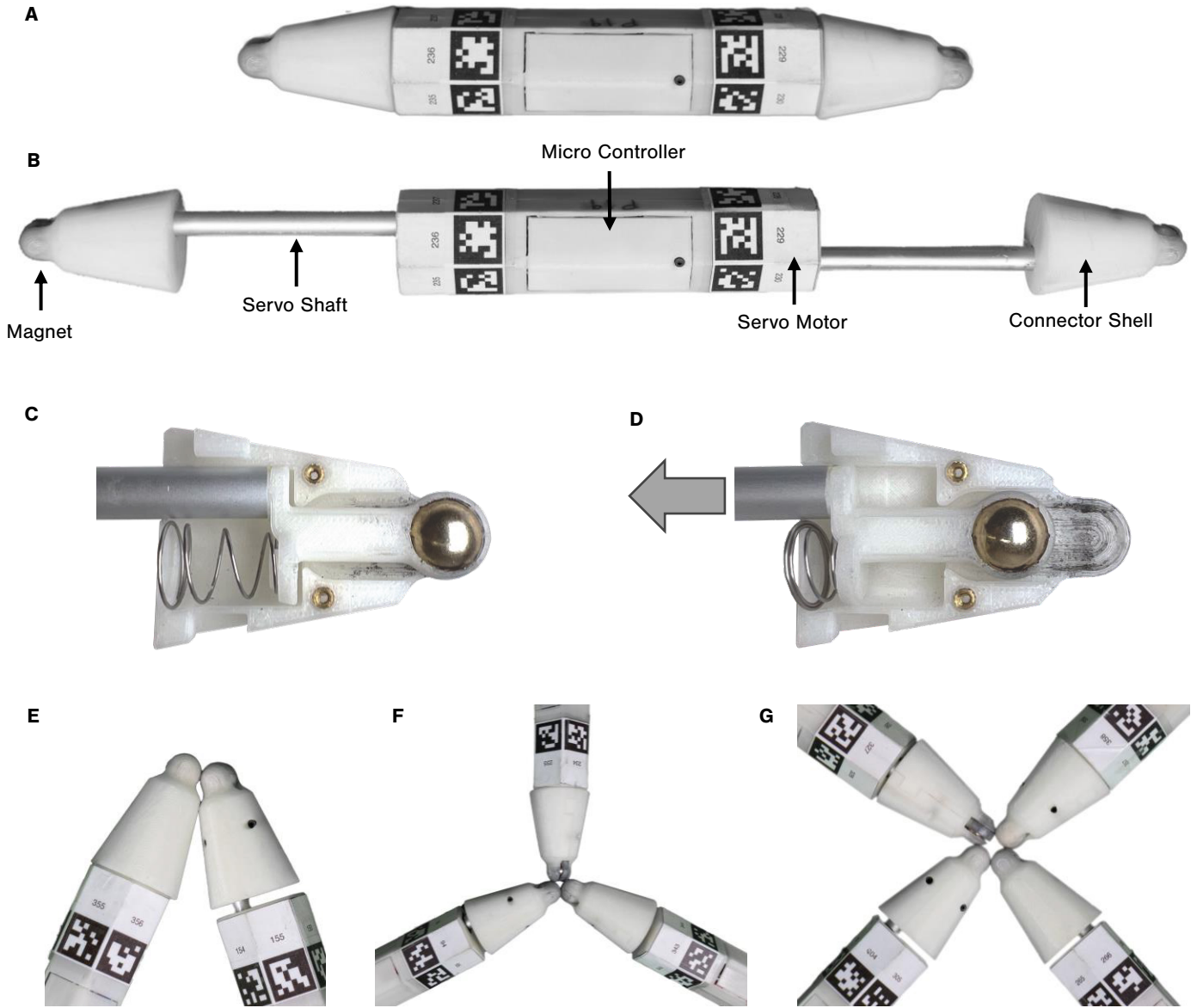

**Figure 2: Truss Link details.** (A) A contracted Truss Link is 28cm long and weighs 280g (B). When fully expanded, a Truss Link can increase its length by over 53% to 43cm. Images (C) and (D) show the interior of the magnet connector in an active state with the magnet exposed at the tip and a fully-contracted, i.e., non-active state with the magnet retracted, respectively. The conical compression spring inside the connector resets the magnet connector to the active state after retracting it, so the Truss Link is ready to connect again. The spherical neodymium magnet is held in position by a magnet holder. The magnet holder allows the magnet to rotate freely to rotate to an equilibrium position when approached by another magnet. This mechanism ensures a strong connection between multiple links from a wide and continuous range of angles. We show connections between (E) two, (F) three, and (G) four connectors.

As truss robots, Truss Links form "scaffold-type" structures and have expanding and contracting prismatic joints (see Fig 2-A and B) rather than rotational ones as they are found in popular cubic-shaped models(*11*). Spherical and cubic robot models have the drawback of forming dense structures, making assembling large robots difficult. Recent developments in modular robotics have shown increased interest in both truss-style and free-form modular robots. Spinos et. al. and Park et al. introduced the first truss-robot capable of self-reconfiguration (*14–16*). Prior truss modular robots such as Morpho and Odin were limited by their attachment mechanism from self-reconfiguring (*17–19*). Both of these systems required connector cubes to join modules.

Many of the well-known cubical modular robot designs, such as Molecubes, M-Blocks, and Smores-EP, had power-sharing or communication channels built into their connectors and, as a result, were limited to a discrete set of attachment angles (*20–23*). Free-form modular robots such as the spherical FreeBot and FreeSN changed this by excluding electronic contacts from their connector; instead, they used a simple magnetic connector with infinite attachment possibilities(*24, 25*). We chose a free-form style connector design to allow Truss Links to effectively self-assemble (See Fig. 2-C through G).

By combining free-form connectors with a truss-style module design, we created a self-assembling platform that forms sparse lattices rather than dense structures. The Truss Link's free-form magnetic connector allows it to attach freely from a wide range of angles without requiring precise alignment. The self-aligning magnet sphere allows multiple connectors to attach to each other, as shown in Fig. 2-E through G. In our experiments, we successfully operated topologies that had up to four connector connections, and we manually assembled topologies with up to six connector connections. The mass of a Truss Link robot scales linearly with the number of Truss Links, while the pull-away force between connectors scales at a lower rate. Thus, 3D structures with more than four connectors connected at a point are more prone to failure. Despite these limitations, the Truss Link is the first truss-style modular robot capable of self-assembly and self-reconfiguration.

## Results

Our results demonstrate that it is possible to form machines that can grow physically and become more capable within their lifetime by consuming and recycling material from their immediate surroundings and other machines. While these results are still nascent, they suggest a step towards a future where robots can grow, self-repair, and adapt instead of being purpose-built with the vain hope of anticipating all use cases. Robot platforms capable of *robot metabolism* open the door to the development of machines that can simulate their own physical development to achieve an objective and then execute that physical development. By acting as open systems, robots capable of robot metabolism bear the potential of forming self-sustaining robot ecologies that can grow, adapt, and sustain themselves, given a continued supply of robot material.

The Truss Link is the first modular truss robot capable of *robot metabolism*. To start, we demonstrate the Truss Link's capacity for self-assembly from individual parts—forming a three-pointed star and a triangle—and by integrating existing sub-structures—forming a diamond-with-tail from a triangle and a three-pointed star. Second, we quantify the probability of random topology formation in simulation given similar randomized initial conditions used in our physical demonstration. Third, we show how Truss Link structures can recover their morphology after separation due to impact via self-reconfiguration or self-reassembly. Fourth, we introduce a way for a ratchet tetrahedron morphology to shed a "dead" Truss Link and replace it by picking up and integrating a found link. Finally, we expand beyond the individual robot and demonstrate how a ratchet tetrahedron robot can assist a 2D arrangement of links to form a tetrahedron.

The Truss Links were operator-controlled in all physical Truss Link experiments using a custom keyboard interface. The interface allows the operator to send commands to selected Truss Links or trigger pre-programmed open-loop control scripts. The pre-programmed controllers allowed us to topple tetrahedrons or make ratchet tetrahedrons and tetrahedrons crawl.

**Multi-stage robot development**

The multi-stage robot development experiment tested whether a 3D structure capable of absorbing and integrating more material could be formed from independent 1D robotic building blocks. If possible, this would lay the foundation for truss robots capable of growing in complexity due to self-assembly and physical development. Next, we quantified the probability of our robotic building blocks randomly assembling into the topologies shown in the multi-stage robot development experiment in simulation. These probabilities provide a reference for the likelihood of achieving these developmental transitions.

      To test our hypothesis, we investigated which environmental conditions facilitated self-assembly. In nature, we see environmental factors crucial to successful development, with early-stage development being most sensitive to environmental conditions. Bird embryos require a hermetically sealed egg to grow, while mammals require a temperature-stabilized womb. Similarly, Truss Links' ability to develop and form new structures is influenced by environmental factors. Identifying a suitable environment was crucial for achieving robot development from basic parts.

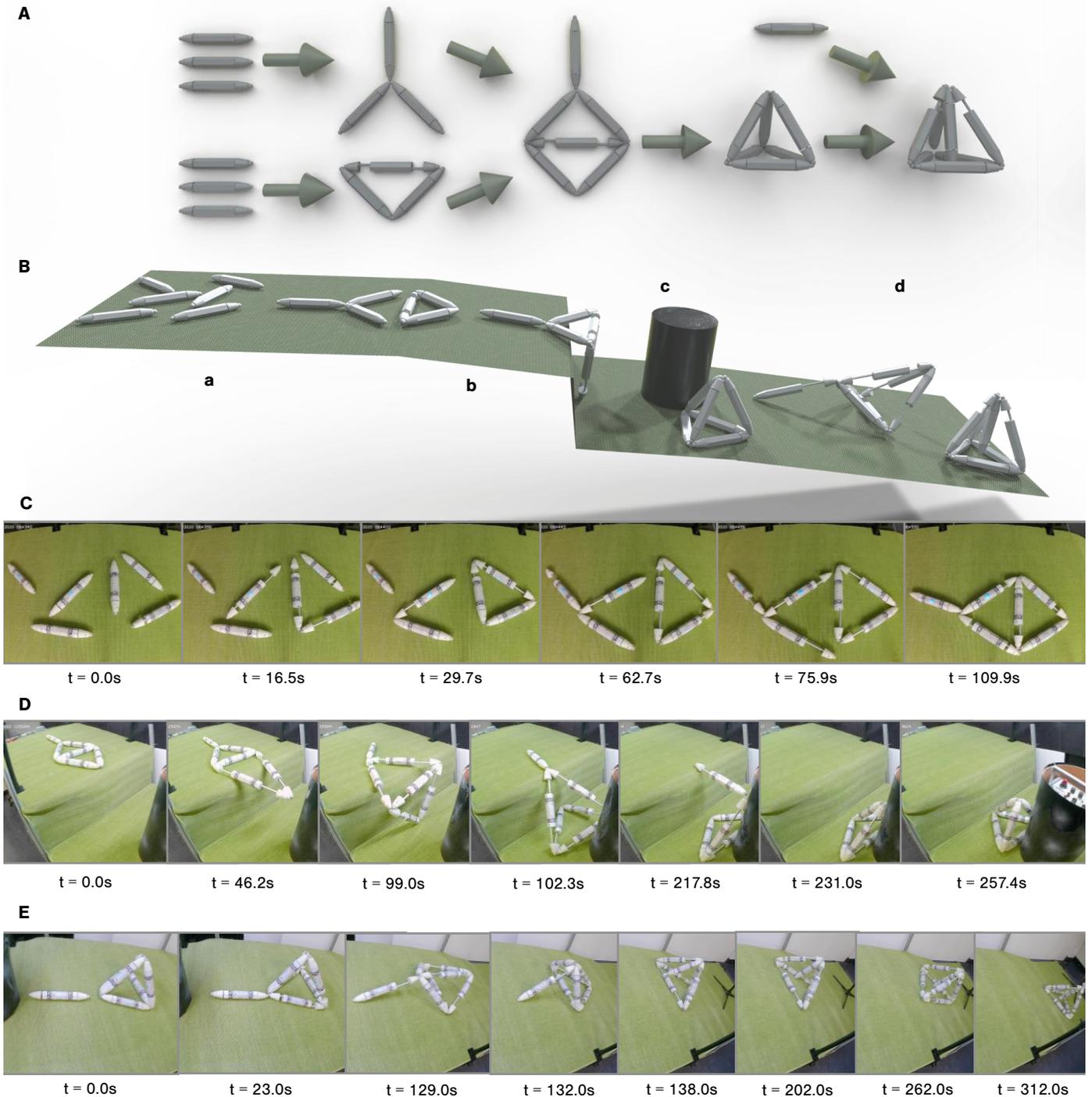

**Figure 3: Multi-stage robot development.** Individual Truss Links can connect to other links to form multi-link robots. These robots can then absorb more links to change their topology. (A) shows a series of topological transitions, starting on the left from a group of individual links and ending on the right with a ratchet-tetrahedron topology. Starting from six independent links, three links combine to form a 3-pointed star shape, and the other three combine to form a triangle. Next, the triangle absorbs the 3-pointed star by connecting to it and becomes a diamond-with-tail topology. The diamond-with-tail then folds itself into a tetrahedron. Next, the tetrahedron finds and integrates a free Truss Link by connecting and picking it up from the ground to form a ratchet tetrahedron. (B) shows the profile view of the experiment environment (not to scale), clarifying where each transition shown in (C-E) took place with section labels (B-a) through (B-d) as a reference. The frame sequences in (C), (D), and (E) show the formation of a diamond-with-tail, a tetrahedron robot, and a ratchet tetrahedron, respectively.

In our simulation environment, we explored what type of world environment would allow us to transition from 2D robot structures to 3D robot structures, particularly the diamond-with-tail to tetrahedron transition. Through experimentation, we found that this transition is more likely to succeed if a diamond-with-tail crawls off a ledge (see Fig. 3 ledge between B-b and B-c), and has an obstacle to lean up against (see black vertical obstacle in Fig. 3-B-c) while folding in on itself—connecting the tail of the diamond-with-tail to its tip (see t=217s-231s in Fig. 3 D). Once we identified a suitable environment, we then built a four-stage (see Fig. 3-B and Fig. S1), 3.9m long and 0.9m wide experiment environment, mimicking the simulated environment. To enable the diamond-with-tail to tetrahedron transition, a ledge followed by an obstacle was placed between stages 3 and 4 (see Fig. 3 B-b and B-c).

The experiment involves a total of seven Truss Links. Six Truss Links start on the first stage (Fig. 3 B-a), and the seventh Truss Link is waiting to be picked up by the tetrahedron on the 3rd stage (Fig. 3 B-c). Throughout the experiment, there are five topological transitions. First, the formation of a triangle and a 3-pointed star from six individual links, followed by the triangle absorbing the 3-pointed star to form a diamond-with-tail (see Fig. 3 B-a). Next, the diamond-with-tail forms by crawling off a ledge and folding in on itself (see Fig. 3 B-b). Last, similar to the tetrahedral mechanism discovered by Lipson and Pollack in (26), the tetrahedron transitions into a tetrahedron ratchet configuration by picking up a found Truss Link and using it as a walking stick (see Fig. 3 B-c).

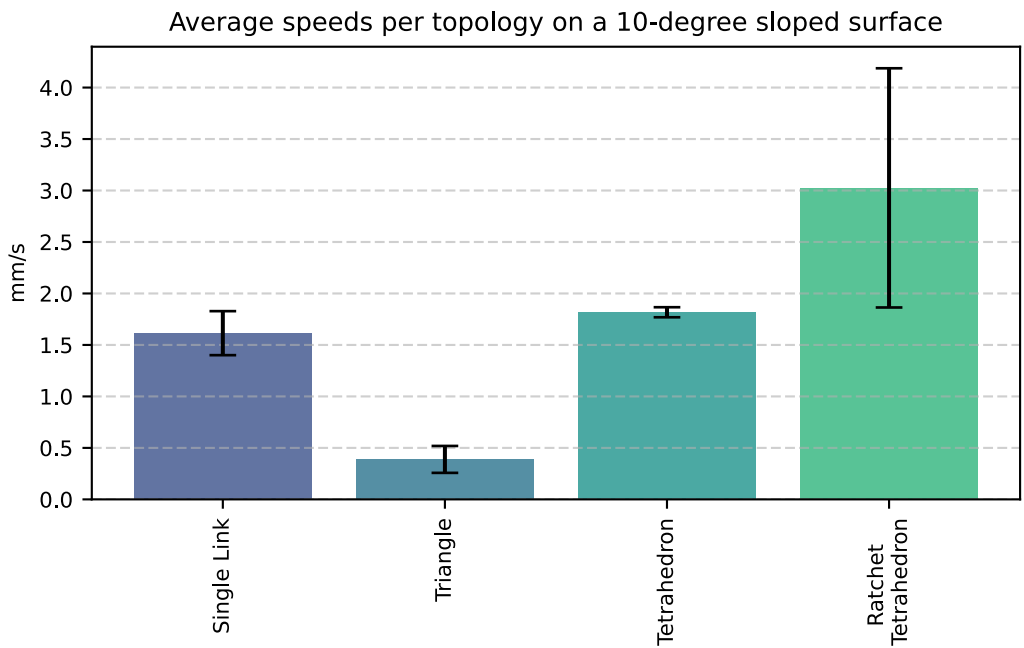

**Figure 4:** Locomotion speed comparison between a single Truss Link, a triangle, a tetrahedron, and a ratchet tetrahedron. The error bars show the standard deviation from the mean. The experiment was conducted on a flat, 10-degree decline.

Each transition in this experiment is designed to produce a more capable topology. Individual links can only crawl forward and backward in 1D space. A triangle and a 3-pointed star can both navigate in 2D space and, therefore, can circumvent obstacles that a single Truss Link couldn't. In contrast with a triangle or a 3-pointed star, a diamond-with-tail can overcome a 25mm tall ledge and can fold itself into a tetrahedron. A tetrahedron can move in 3-dimensions by toppling onto obstacles that were inaccessible to previous topologies. A ratchet tetrahedron increases its walking speed by over **66.5%** on a 10-degree slope compared to a tetrahedron (see Fig. 4). See supplementary videos S2 and S3. After assessing its feasibility in simulation, we successfully reproduced every transition of the experiment on the physical platform. Our experiments demonstrated that three independent links can combine to form a triangle and a 3-pointed star configuration. Next, we showed that a triangle can connect to and integrate a 3-pointed star to form a diamond-with-tail shape that can further fold itself into a tetrahedron. Finally, we demonstrate how a tetrahedron robot can consume a found Truss Link and integrate it into a tetrahedron-ratchet configuration (see Fig. 3-E).

**Simulated vs. real-world crawling speeds**

We compared the crawling performance between topologies: individual Truss Link, triangle, tetrahedron, and ratchet tetrahedron. We replicated the 10-degree slope environment and the gates used on the physical robot in our PyBullet simulation. We normalized the speeds of all topologies by their body length: 28cm for the individual Truss Link and 24.5cm for the triangle, tetrahedron, and ratchet tetrahedron topologies. Further, we multiplied their body length per second speed by the time each topology takes to execute one crawling cycle, i.e., "take a step." The cycle time is 16 seconds for the ratchet tetrahedron and 36 seconds for all other topologies. The physical friction conditions of the Truss Links on the carpet cannot be accurately replicated in the PyBullet simulation environment. The simulated ramp acts as a hard, untextured surface with a lateral friction coefficient of 0.89 and spinning and rolling friction coefficients of 0.02 and 0.003, respectively. Please see our code repository for the corresponding simulation scripts to replicate these simulations.

We filtered the simulation data using a z-height threshold to exclude periods where the topology may have moved off the platform. Each data point corresponds to a crawl cycle. We estimated locomotion speeds throughout the experiment by fitting a linear regression line to the y-position and timestep data over four consecutive cycles, advancing the window by two cycles at each step. Then, we averaged the resulting speeds and computed their standard deviations. Finally, we normalized all values to body lengths per cycle.

**Table 1:**

**Simulated vs. real single-direction locomotion speeds on a 10-degree downward slope in body lengths per cycle**

|  | Single Link | Triangle | Tetrahedron | Ratchet Tetrahedron |
|---|---|---|---|---|
| Simulated | 0.3554±0.0014 | 0.3382±0.0446 | 0.3274±0.0159 | 0.3867±0.2397 |
| Real | 0.2070±0.0275 | 0.0573±0.0168 | 0.2674±0.0063 | 0.1979±0.1494 |

The results from the simulated runs for each topology, as well as the average body length normalized speeds of our physical experiments and their corresponding standard deviation, are shown in Table 1. We find that the crawling speeds between topologies vary less in simulation. Most notably, the simulated triangle crawled faster than expected, while the simulated tetrahedron crawled slightly slower.

We observed that the physical Truss Links showed more backsliding behavior, especially the triangle. On the consistent and smooth surface in the simulation environment, the crawling speed of the triangle was slightly faster than that of the tetrahedron, while in our physical experiments, it was significantly slower. The weight shifting performed by the tetrahedron by moving its upper three links does not result in the same benefit on the hard simulated surface as it did on the soft carpet where the connector's edges would sink in. The simulated links don't suffer from manufacturing errors: slightly rotated motor shafts and rough edges on the Truss Link body that could result in added friction. The real triangle's crawling behavior was more brittle than that of the tetrahedron: it was more likely to get stuck due to a loose servo shaft or to rotate due to differences in friction between the two front Truss Links.

All topologies were faster in simulation than in the real world, which likely is a direct result of the difference in friction and surface texture. The ratchet tetrahedron benefitted most from the simulated friction condition and is faster than all other topologies on average, but highly volatile in its crawling speed. The ratchet tetrahedron in simulation and on the physical robot has a propensity to rotate along its vertical axis and deviate from its intended path. Given its instability, the ratchet tetrahedron crawling behavior doesn't lend itself well to open-loop operation. In contrast, the individual Truss Link and the Tetrahedron demonstrated the most consistent crawling behavior in simulation with a smaller standard deviation than the other two topologies.

**Simulated Morphology Formation Probabilities**

Teleoperated, physical experiments don't highlight the difficulty of forming different morphologies. Thus, we quantified the formation probabilities of different morphologies in simulation. We simulated the morphological development experiment environment, spawned the Truss Links randomly in the same section of the experiment environment as the physical experiment, and randomized the control inputs. We added walls to the simulated experiment environment to prevent Truss Links from falling off. To track the morphologies during simulation, we hashed all magnets based on their x and y locations into a 2D occupancy grid with 16mm-by-16mm square cells and then considered all magnets that were within the same cell or within neighboring cells to be connected. Based on our empirical observation of the physical platform, this is a reasonable assumption since two magnets within that range would inevitably snap together. Next, we represented the morphology as a graph by treating the links as edges and groups of connected connectors as nodes. Finally, we computed the Weisfeiler-Lehman hash for each graph representing a specific morphology (see supplementary materials S1.7). The resulting formation probabilities provide a numerical reference for the likelihood of the transitions in our previous experiment occurring by chance without human assistance.

| Formation | Probability | Formation | Probability |
|---|---|---|---|
| 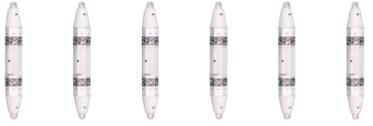 | 100% | 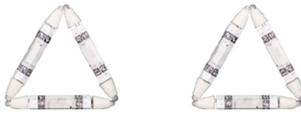 | 8.4% |
| 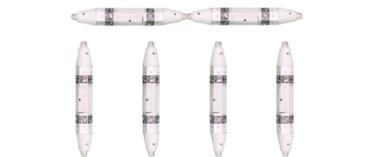 | 98.6% | 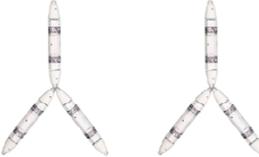 | 64.35% |
| 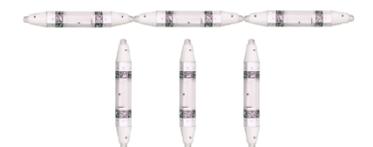 | 97.6% | 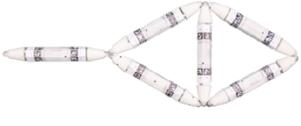 | 44.3% |
| 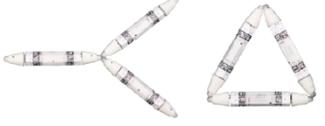 | 9.2% | 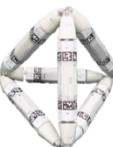 | 0% |

**Figure 5:** Simulated random topology formation probabilities over 2000 20-minute simulation runs

The analysis was conducted on 2000 random experiment runs, each limited to 20 minutes of simulation time. Experiments that were initialized with links already connected were excluded from the analysis and not counted towards the 2000 analyzed experiment runs. For each run, we stored the set of all morphologies that occurred during the simulation. From this data, we extracted the probabilities shown in Fig. 5.

The formation probabilities show that some but not all of the morphologies could be reproduced spontaneously from the random initial state with random motor commands within 2000 attempts. It becomes apparent that the formation of a diamond-with-a-tail is highly likely from the spawn locations chosen in the experiment, given that it occurred in 44.3% of the experiment runs. This high probability points towards an initialization bias, which was intentional since the initialization was supposed to mimic the one used in the physical experiment. However, it is worth noting that just 9.2% of the experiment runs exhibited a three-pointed star and a triangle simultaneously, indicating that most diamond-with-tail shapes were not formed as demonstrated on the physical robot by combining a triangle with a three-pointed star.

From the physical experiment, we learned that forming the tetrahedron and the ratchet tetrahedron is possible but challenging without an added controller. The tail-link of the diamond-with-tail shape is only connected at one point and thereby position-constrained but free to rotate around the connection point. Similar to an inverted pendulum, this was challenging for a human operator to learn. Over the course of 37 attempts, the environment was adjusted: for example, adjusting the distance and tilt of the cylindrical obstacle and flattening the carpet. At the same time, the human operator had to learn to control the robot using the keyboard interface (see supplementary materials Fig. S4). After the last environment adjustment, the first successful tetrahedron was formed on the sixth attempt. Thus, we can conclude that more randomized runs and more simulation time would have produced a non-zero probability for the tetrahedron. Following this line of reasoning, Truss Links could "grow" on their own even if they acted randomly.

**Damage recovery**

Biological life's ability to self-heal by reforming broken bonds or growing back parts inspired us to attempt robot self-repair by reforming broken bonds between Truss Links. The magnetic connections between Truss Link connectors form pre-determined breaking points, reducing the risk of physical damage to the Truss Link hardware from impact. In this section, we explore how this feature enables robots to recover their original topology after being separated upon impact. In our tests, we let triangle, three-pointed star, and diamond-with-tail robots crawl off the 30cm tall ledge between stage B-b and B-c of the experiment setup shown in Fig. 3, such that they disconnected on impact and then attempted to regain their original morphology.

For this experiment, we limited damage to a loss of the original topology due to broken connections between Truss Links. This is in contrast to the breaking or malfunctioning of Truss Links. We intentionally kept the drop height low to avoid damage to the Truss Links. In the case of a broken Truss Link the robot would need to get rid of the broken part and replace it with a functioning one, as shown in our next experiment.

The violent disconnections after impact and the slopes of the experiment environment resulted in hard-to-predict outcomes that were difficult to control for the operator. Thus, several reconstruction attempts were not successful. We share examples of successful shape recovery for all three topologies below.

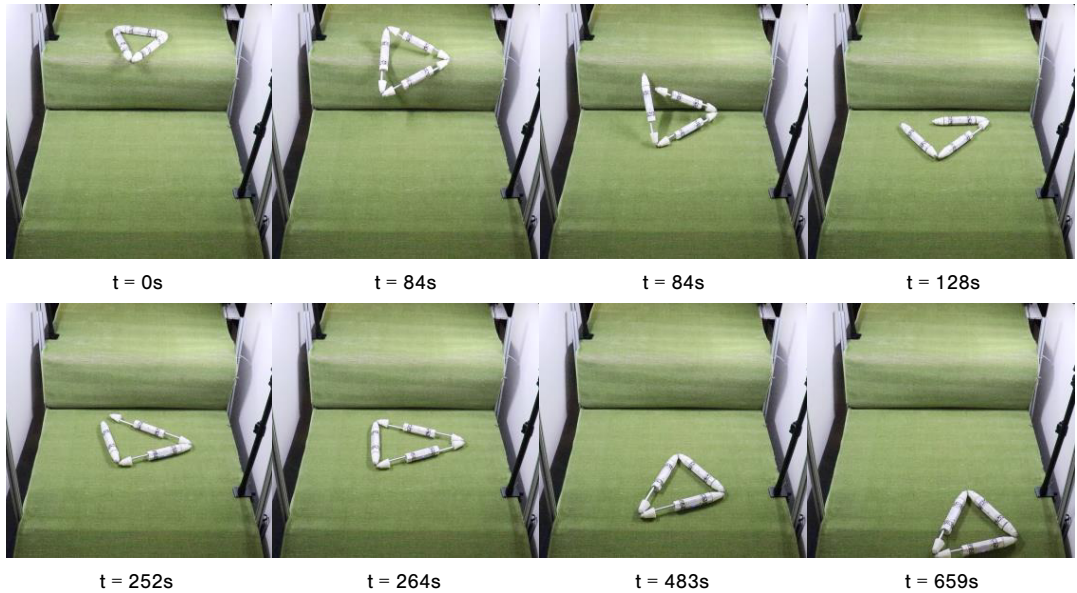

**Figure 6: Triangle shape recovery**. A Truss Link triangle robot crawls off the ledge, breaks a connection due to the impact, proceeds to recover its triangle shape, and crawls away.

The triangle is a fully constrained shape and, therefore, a naturally stable planar topology. As a result, the triangle resisted breaking any connections on several attempts. All triangle connections are strong two-connector connections without unconstrained degrees of freedom. If one connection did break, the other two would usually hold.

In one failed attempt, the triangle managed to break both connections with the back Truss Link, which, due to the sloped surface, rolled and re-connected at the connection point between the other two links—forming a three-pointed star. In another attempt, the back Truss Link broke a single connection, but the operator didn't manage to re-form the triangle within the bounds of the filming setup and thus aborted the attempt.

A successful damage recovery sequence is shown in Fig. 6. Notice how at t=84s the triangle's back left connection gets disconnected due to the asymmetric fall. After extending its back Truss Link, the triangle recovers the connection by extending its front-right Truss Link.

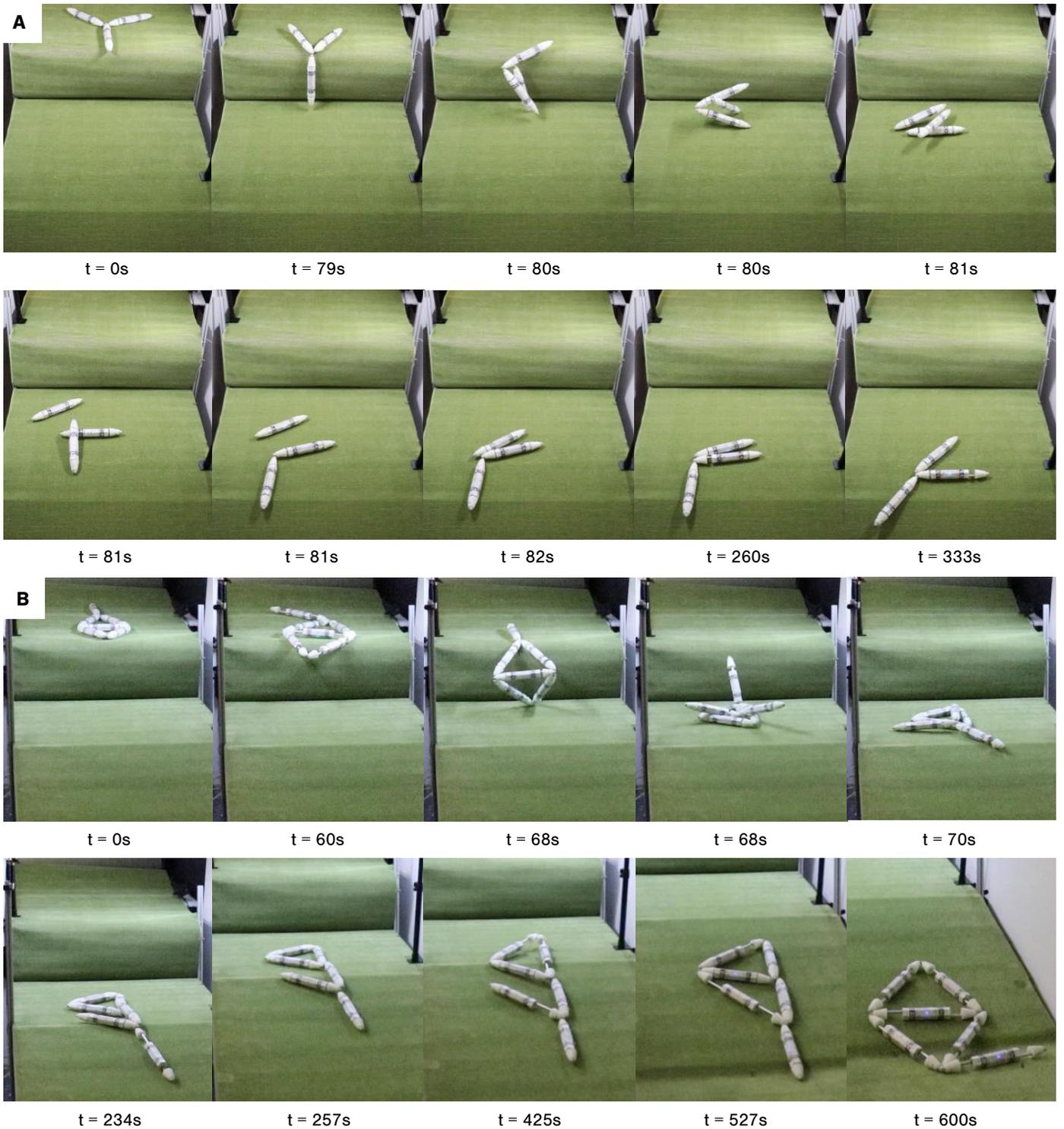

**Figure 7: Three-pointed star and diamond-with-tail shape recovery**. (A) A three-pointed star robot crawls off a ledge and breaks all Truss Link connections. The robot then regains a three-pointed star shape and crawls away. (B) A Truss Link diamond-with-tail robot crawls off a ledge and separates into a three-pointed star and a triangle robot. The three-pointed star robot lands on top of the triangle robot. Next, the three-pointed star robot crawls off the triangle and reconnects to it, ultimately regaining the diamond-with-tail shape.

In contrast to the triangle, the three-pointed star topology is under-constrained: all three links are only connected on one end. As a result, it is less predictable, harder to control, and more brittle. Several attempts failed spectacularly, with links being flung down the ramp or rolling away, thereby making shape recovery impossible.

A successful sequence showing the damage recovery of a three-pointed star is shown in Fig. 7-A. At t=80s the three-pointed star drops and completely disconnects following the impact. The three-pointed star recovered its original form after the links rolled near each other, and the Truss Link facing in the 2 o'clock direction at t=82s shuffled itself to a 3 o'clock position (t=260s). The three-pointed star was able to crawl away after recovering its shape.

To assess if a larger structure could recover its original shape, we conducted the experiment using the diamond-with-tail topology consisting of six links. Only one of the connections on the diamond-with-tail are two-connector connections; the other three connections are three-connector connections.

The diamond with tail structure is under-constrained and similarly unstable when falling as the three-pointed star. When falling off the ledge, the front of the structure crashes into the experiment surface, while the back end is still sliding or falling, adding additional force to the Truss Link connections and breaking them. As a result, the Truss Links further back in the structure can fall on top of the links in the front.

A successful recovery sequence of a diamond-with-tail that separated in the middle into a triangle and a three-pointed star is shown in Fig 7-B. The three-pointed star landed on top of the triangle and had to shuffle itself off of the triangle first before reconnecting. The three-pointed star managed to connect to the lower-right vertex of the triangle. After four minutes of maneuvering, the second Truss Link of the three-pointed star reconnected to the lower-left corner of the triangle. Finally, the reformed diamond-with-tail moved itself off the ramp.

**Replacing a "dead" Truss Link**

Truss Link structures can self-assemble, but can they self-repair? In this experiment, we tested if a ratchet tetrahedron could recover from losing its ratchet Truss Link due to power loss. Truss Links are programmed to fully contract and detach by retracting the magnets inside the connectors once battery power drops below a critical threshold. Thus, similar to apoptosis in multicellular organisms (i.e., programmed cell death), the robot can shed a Truss Link that is no longer needed or threatens the robot's overall functionality.

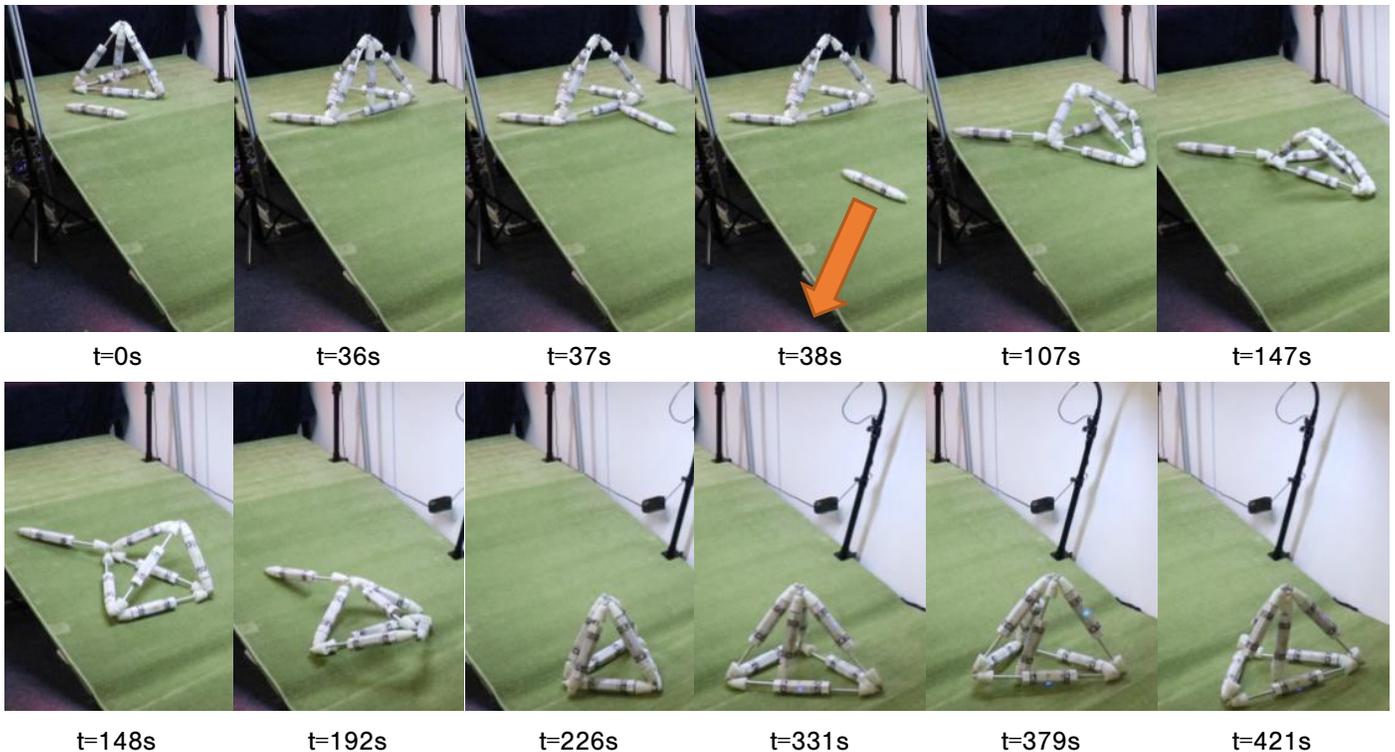

**Figure 8: Ratchet-tetrahedron sheds "dead" ratchet Truss Link and picks up a replacement.** The ratchet tetrahedron approaches the single Truss Link and latches onto it. Next, it sheds the dead link: the fully contracted and detached "dead" Truss Link falls off of the tetrahedron and rolls down the slope. The tetrahedron then topples itself twice to re-orient itself to pick up the newly found Truss Link. After the pickup at t=192s, the tetrahedron swings the Truss Link into its center and ratchets away.

In the frame sequence shown in Fig. 8, the ratchet tetrahedron first finds and connects to a replacement Truss Link with its right-front-bottom vertex. Next, as shown in Fig. 8 at t=36s and following, the ratchet Truss Link is triggered to execute its death sequence, where the Truss Link fully contracts both servos. This causes the ratchet Truss Link to let go of its connection and then roll away due to the environment slope at t=38s. Next, the tetrahedron topples first forward (see t=107s) and then to the right (see t=147s) to get into position to pick up the replacement Truss Link. Finally, at t=192s, the tetrahedron picks up the new Truss Link, swings it inside itself (t=226s to 331s), and then continues to use it as a ratchet at t=379s and following.

This experiment was conducted on stages three and four of the experiment (see Fig. 3-B c and d). The experiment environment has a slope that is necessary to enable the tetrahedron to pick up the found Truss Link. The slope also has the benefit of allowing a shed Truss Link to potentially roll away and thereby not interfere with the process of picking up the replacement link.

**Robot-to-robot assisted reconfiguration**

Earlier, we have demonstrated how a ratchet-tetrahedron can be assembled from individual Truss Links. However, the transformation from diamond-with-tail to tetrahedron, as shown in Fig. 3, is not trivial and requires specific environmental conditions. Here, we study if, once a tetrahedron has been formed, the transition from a two-dimensional flat pattern to a tetrahedron could be facilitated by robots assisting each other.

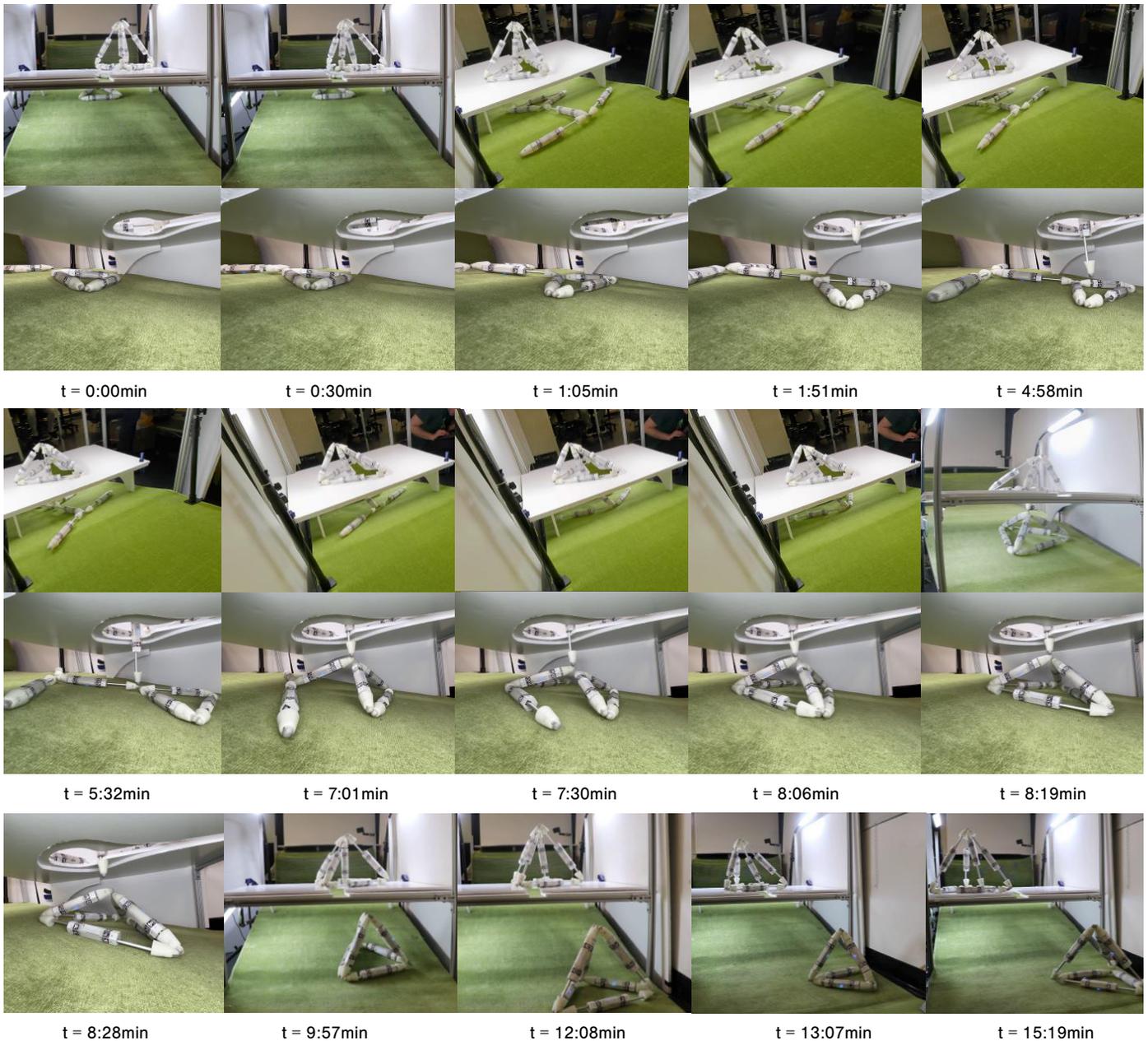

**Figure 9: Robot-to-robot assisted tetrahedron formation**. A ratchet tetrahedron uses its ratchet Truss Link to fish through a hole in the white platform for the vertex where the triangle and the three-pointed star are connected. After being lifted up, the three-pointed star connects to the two free vertices of the triangle, forming the tetrahedron. The different, time-synchronized camera angles in the frame sequence were picked based on which camera provided the most informative view of each stage.

In this experiment, we identified a way to erect multiple consecutive flat patterns into tetrahedrons one after another, thereby significantly lowering the difficulty of forming more tetrahedrons after the first ratchet tetrahedron is formed. Inspired by the teardrop-shaped canyon cross-sections found in Leprechaun Canyon, the experiment environment features a raised platform with a narrow opening and a sloped surface below. From this elevated position, the ratchet tetrahedron can assist other links to extend into the third dimension.

A frame sequence of this experiment, including multiple camera angles, is shown in Fig. 9. A ratchet tetrahedron can position itself on the raised platform above the opening (see t=0min to 1:51min). The raised platform mimics a washed-out canyon with overhanging walls that only leave a narrow gap at the top. A three-pointed star and a triangle then crawl underneath it. The three-pointed star connects to one of the triangle vertices by extending one of its links, as shown at t=0:30min. The ratchet tetrahedron can then reach down through the narrow gap—like a crane, connect to that same vertex and lift it up (see t=4:58min to 8:28min). Since the whole weight of another tetrahedron exceeds the holding power of the ratchet link's magnet connection, it has to support the ratchet Truss Link body on the edge of the gap (see t=7:01min). In this way, the links below can move around without risking the structural integrity of the ratchet above. The three-pointed star's two free links then shuffle their way toward the triangle's vertices, as shown from t=7:01min to t=8:06min, until they connect—voila, a tetrahedron is formed. Next, the ratchet tetrahedron needs to disconnect from the tetrahedron. The ratchet tetrahedron drops the newly formed tetrahedron by fully contracting one side of its ratchet Truss Link and retracting the magnet inside the connector (see t=8:28min). Then, at t=9:57min, the newly formed tetrahedron crawls away. At this point, the next three-pointed star and triangle could come along and undergo the same assisted transformation.

We empirically explored various experiment setups and found over the course of 61 trials that a platform with a slot rather than a hole and a three-pointed star connected with one link to the triangle as the flat pattern facilitates the transformation. Using a diamond-with-tail topology as the starting point for the transformation was unsuccessful. After the last change to the experiment setup was made, three tetrahedrons were formed over ten attempts. Common causes of failure were operator error leading to the ratchet tetrahedron collapsing and Truss Link malfunctions due to low-battery or WIFI connectivity issues.

Through this experiment, we showed that the difficulty of forming a Truss Link tetrahedron can be reduced by robots assisting robots. This method of tetrahedron formation could be repeated without navigating the risks of folding an under constrained three-pointed star by crawling it off a drop. Finally, the transformation shown in this experiment aligns with the constraints of the *robot metabolism* and shows that robot development need not be a solitary endeavor.

## Discussion and conclusion

We presented a robotic system that can produce structures that can develop physically, i.e., grow in size and capability, by absorbing and integrating found Truss Links or existing Truss Link structures. Many self-reconfiguring robotics systems have been demonstrated in the past, including our own systems capable of self-production (7). Unlike the Truss Link platform presented here, none of these systems could develop from single 1D cells to a full 3D robot, while systematically improving its own capability in the process and without requiring external machinery.

## Limitations and future work

The robot structures presented in this paper are very simple. This is a direct result of the still nascent stage of the field of self-reconfiguring modular robotics and the software infrastructure surrounding it. The Truss Link's design was deliberately kept to the bare minimum required for this demonstration. We believe that smaller and simpler building blocks will ultimately span a larger space of potential robot morphologies. However, practical considerations dictated by available linear actuators limit the expansion ratio, weight, and strength of each link in this study. In future work, we aim to develop microscale models that would allow the construction of single robots composed of millions of cells.

High cost and manufacturability, sensor integration, communication and control, and simulation are known challenges for modular robotic systems. Our high-fidelity simulation was sufficient to explore *robot*

*metabolism* as a proof of concept, but it lacked the performance for machine learning-based control algorithms. A massively parallel, high-fidelity simulator would open the door to studying both design exploration and validation, sensor integration, and communication and control for the next generation of modular robots capable of *robot metabolism*. Thus, we plan to develop such a simulation environment for Truss Links in future work.

The Truss Link platform was designed around off-the-shelf components and built using commonly accessible tools to make it easy to replicate by anyone. With more than $200 in material cost per unit, Truss Links are neither cheap nor designed for mass production. A custom actuator design combined with a custom circuit board and a single battery power source could decrease the form factor significantly while still maintaining a high expansion ratio, enabling more impressive self-assembly results. In addition, a custom circuit could integrate current sensing for the actuators, encoders, an IMU, power management, and a charging circuit. This type of custom hardware, while promising more impressive results, would increase the cost of a single unit and make the research harder to replicate. We believe the promise of low-cost, mass-manufactured modular robots can only be achieved once academic research demonstrates a business case with a clear path to profitability for this technology and thereby sparks industry adoption.

Integrating sensors into the modules comes with the challenge of communicating and processing the sensor data. Sensors such as IMUs, magnetometers, current sensors, cameras, microphones, etc., can produce high-frequency, high-resolution time-series data that would have to be processed onboard using an edge computing module instead of a simple microcontroller. Given this type of system, Truss Link modules could then be programmed to modulate their behavior in a decentralized fashion based on sensor readings while following a global objective.

Since Truss Link structures can change their topology, the controller must deal with changing kinematics and dynamics, as well as underactuated joints. First, we plan to explore a centralized control solution assuming perfect pose information of all Truss Links. Using search algorithms combined with reinforcement learning, we can identify robot topologies and learn corresponding locomotion controllers. In addition, we can learn transition controllers that allow the robot to morph from one topology to another. Second, we plan to explore a decentralized control approach where each Truss Link's behavior is dependent on its sensor readings, local signaling between neighboring modules, and a globally shared objective. For example, by fusing force readings from actuators with IMU, and 360 camera readings, Truss Links could learn an end-to-end controller for self-assembly. In addition, Truss Links could be equipped with sonar, radio frequency, or light-based local signaling equipment as pathways for learned communication patterns. Ultimately, transferring these learned controllers from simulation to reality poses a significant challenge due to unaccounted-for differences between the simulated Truss Links and the physical system. For these reasons, we see the development of a high-fidelity, massively parallel simulation as the logical next step.

Applications for platforms capable of robot metabolism are distant but inevitable. As our economic welfare grows increasingly dependent on robots, it becomes necessary that these robots can take care of themselves physically. Given their increasing complexity, we see it as unlikely that human engineers will be able to maintain the growing numbers of robotic systems or manually adapt them to new tasks and environments in the long term. We must understand how to build building blocks that enable self-sufficient robots to adapt, self-sustain, and grow. In essence, we need to create a self-sustaining robot ecology.

## Materials and Methods

Here, we share additional information on our experiment environment and how the walking speeds of different Truss Link topologies were measured. We provide in-depth information on the Truss Link's hardware and design. We explain how the Truss Links were coordinated using our Truss Link server and controller. Next, we address the key aspects of our customized PyBullet simulation environment used in our experiments. A rendered video of a diamond-with-tail forming in a randomized experiment can be seen in supplementary video S4.

**Experiment Environment Details**

Our experiment setup (shown in Fig. S1) was designed to allow the Truss Links to transition from single links to a ratchet tetrahedron. The experiment environment was designed with adjustable slopes for each stage. We initially set the slopes to the values that were used in the simulation and then adjusted the slopes as needed to achieve the transformations shown in the multi-stage robot self-assembly experiment. Stages one to four are 1.2m, 0.6m, 0.6m, and 1.2m long, respectively. The surface is built from 6mm thick plywood that is covered with a layer of 10mm thick foam board to smoothen the stage transitions from stages one to two and three to four. Additionally, a cardboard cylinder containing weight was placed as an obstacle on stage three to allow the diamond-with-tail to fold itself into a tetrahedron in a controlled manner.

All our physical experiments were conducted on a 4mm-pile Polypropylene carpet to ensure a consistent experiment surface. We used stationary cameras and LED lighting to film each experiment.

**Walking speed experiments**

To assess the walking speed of different topologies during successful crawl cycles, we conducted a repeated locomotion experiment and plotted the results. Truss Links rely on differential friction for crawling and can get slowed down or stuck on uneven surfaces. Since this experiment aimed to assess speeds during successful walking or crawling maneuvers, we excluded video sequences where a topology got stuck on an uneven surface or crawled outside of the experiment setup from the measurement data. The speed measurements reported in this section were all collected on a 10-degree downward slope to mimic the conditions of stage four of the experiment setup. The gates used in the speed experiments were manually programmed, tuned based on empirical observations, and then executed in an open-loop fashion. We marked the experiment surface with a line every 5cm to track the robot speeds from the video footage.

The experiment results are shown in Figure 4. The findings show that a crawling link, while only being able to move in a single dimension, is faster than a triangle. The triangle, which is superior to the individual Truss Link by being able to move in two dimensions, underperforms the single Truss Link's speed due to its increased weight and inopportune Truss Link angles. The crawling tetrahedron is slightly faster than a single Truss Link and demonstrates the most consistent performance. The ratchet tetrahedron is the fastest topology tested in this experiment but also the one with the most variance in speed. During its crawling motion, the ratchet tetrahedron tends to rotate and orient itself away from the slope direction, which causes it to slow down or move sideways rather than forwards. This instability in the ratchet tetrahedron gate could be compensated for during closed-loop operation but was included intentionally to reflect the raw system's dynamics.

The raw data and data processing scripts used to create the data for Fig. 4 and Table 1 are included in the *supplementary_data.zip* file. See the supplementary materials for further information.

**Truss Link design**

The Truss Link is the homogenous building block of our truss-type modular robot system. Truss Links allow for the construction of chain and lattice structures. The main hardware innovation is the Truss Link's compliant magnetic connector that passively orients the polarity of a 1.27cm diameter neodymium magnet sphere inside the connector to generate an equilibrium of attraction among all modules trying to connect at a single point. According to our in-line dynamometer pull-away tests, two connectors require a pull-away force of approximately 13.7N to be separated. Modular robot designs commonly incorporate communication channels into their connectors (*27*). We opted not to use the connector for power sharing or communication to reduce the design complexity and increase the connector's versatility. Our design can form connections without needing passive connector blocks, such as the ones used in Morpho or the Odin robot, since that would have complicated self-assembly (*17*, *18*).

We designed the Truss Link platform to form a tetrahedron structure capable of picking up a Truss Link attached to a base vertex by toppling itself over. To achieve this motion, the tetrahedron must be able to sufficiently shift its center of mass without collapsing. A geometric analysis revealed that the Truss Link's minimum expansion ratio—the maximum length of expansion a Truss Link can achieve as a percentage of the minimum length of a link—must be more than 41.5 percent to allow for the tetrahedron toppling behavior. Our current Truss Link design with a contracted length of 28cm and an expanded length of 43cm achieves an expansion ratio of 53%.

Each Truss Link body comprises two prismatic actuators, one Particle Photon microcontroller, a WIFI antenna, a voltage regulator, a voltage divider, and batteries. As our actuator, we chose the 100mm stroke length Actuonix L-12I linear servo with a gearing ratio of 210:1. Its small form factor and simple control interface facilitated integration. Since the motor housing of the linear servo is the same size for each stroke length, we maximized the Truss Link's expansion ratio by picking the Actuonix L-12 servo model with the maximum stroke length.

The two linear actuators can be both independently and jointly actuated. The Truss Links were designed with a passive attachment/detachment mechanism in each connector. Considering the detachment mechanism as a separate degree of freedom (DoF), each Truss Link is a 4-DoF system. Aside from the *Replacing a "dead" Truss Link*, and *Robot-to-robot assisted reconfiguration* experiments, we treat each Truss Link as a 2-DoF system since the attachment/detachment mechanism is not used.

Truss Links are powered by two removable single-cell 380mAh Lithium Polymer batteries that are connected in series. We step down the voltage to 5 Volts for the Particle Photon via a voltage regulator and use a voltage divider to monitor the battery voltage via the onboard 12-bit analog-to-digital converter. Please refer to the Supplementary Materials PDF for further technical details regarding the Truss Link system.

**Truss Link connector**

The Truss Link uses a free-form magnetic connector with a detachment mechanism. The connector comprises FDM-printed body shells and a magnet holder, as well as a 12.7mm diameter N52 neodymium magnet sphere, a conical compression spring, and two screws and two heat-set inserts (See Fig. 2-C). The entire connector is held in place via the magnet holder which is screwed and hot-glued directly into the servo shaft. The magnet holder constrains the magnet's position while allowing it to rotate freely, so it can align its polarity when connecting with other connectors. To reduce friction during magnet alignment, we apply a dry graphite lubricant on the inside of the magnet holder.

The connector detaches by retracting the magnet inside the connector shell, thereby reducing the magnetic field outside the connector. The magnet gets retracted by 1.2 times its diameter to ensure secure detachment. The connectors' shape results from a tradeoff between optimizing magnet connection strength while maintaining the ability to disconnect via the magnet pullback mechanism. The small rim below the connector tip prevents other connectors from simply sliding along the outside of the connector during detachment, while the conical shape helps create distance between the magnet sphere and any outside connectors.

The connector detaches by fully retracting the Truss Link's servo, thereby pushing the connector shell against the center body, and retracting the magnet holder. When expanding after a full retraction, the magnet holder resets itself back into an attachment-ready position using the conical spring inside the connector (See Fig. 2-C and D). To ensure a smooth resetting behavior, we smoothen and grease the friction points between 3D printed parts.

To minimize the center body size, the servos were aligned in parallel but in opposite directions. As a result, the servo shafts are not centered on the Truss Link body's central axis but are placed next to each other. To compensate for this offset, we added an offset to the magnet holder, such that the connector tips are centered on the body's central axis. Centering the connector on the link's central axis allows us to balance the geometry reduce rotations along the link axis during maneuvers, and thereby improving the stability Truss Link structures.

We noticed during experiments that connectors and servo shafts sometimes come loose after repeated use and then rotate out of alignment. One possible solution could be to retrofit the servo motors with square shafts and square shaft guides. This would prevent both the connector shell and shafts from rotating and unscrewing themselves.

## Acknowledgments

**Funding**
We thank the NSF AI Institute in Dynamic Systems, NSF NRI, and DARPA Trades for their support.
DARPA TRADES COLUM 5216104 SPONS GG012620 01 60908 HL2891 20 250
NSF NRI COLUM 5216104 SPONS GG015647 02 60908 HL2891
NSF NIAIR COLUM 5260404 SPONS GG017178 01 60908 HL2891

**Ethics Approval**
Not applicable

**Consent to participate**
Not applicable

**Availability of data and materials:**
All data are available in the main text or supplementary materials.

**Code Availability**
Please visit robotmetabolism.github.io

**Author Contributions**
- Research proposal; HL, PMW
- Truss Link prototype design: PMW
- Truss Link design revision and manufacturing: QAB, JW, PP, DY, MEM, NNG, RTK, PMW
- Communication and control software: PMW, RB., MZ, AS, SK
- Simulation environment: RB, DK, OM, EHT, PMW, SB, YH, PJ
- Physical experiments: PMW, MZ, QAB, RB, JW, SK
- Simulated Experiments: PMW, RB, MZ
- Filming and Photography: QAB
- Data analysis: MZ, RB, PMW
- Guidance: HL, PMW, JG
- Writing—original draft: PMW, HL
- Writing—review & editing: PMW, HL, JG, YH

**Competing interests**
Authors declare that they have no competing interests.

**Supplementary Materials**
Please refer to the supplementary materials document and supplementary video files.

# Supplementary Materials for

## Robot Metabolism: Towards machines that can grow by consuming other machines

Philippe Martin Wyder* *et al.*

*Corresponding author. Email: philippe.wyder@columbia.edu

**This PDF file includes:**

    Supplementary Text
    Figs. S1 to S4
    Table S1
    Movies S1 to S5 captions
    Supplementary Data description

**Other Supplementary Materials for this manuscript include the following:**
    Movies S1 to S4
    Raw experiment data and analysis scripts

**Supplementary Text**

Truss Link server and controller

To coordinate the motion of multiple Truss Links, we wrote a server program that manages the communication with Truss Links and ensures synchronized command execution. Our experiment scripts can then use this server to send motor commands to the physical Truss Links. The server allows us to conduct experiments that treat the Truss Links as independent entities with separate controllers or as multi-link robots.

Since each Truss Link acts as an independent client, the server handles multiple parallel connections, one with each Truss Link. The server uses a multi-threaded socket handling approach. Once started, the server software runs a *Listening Thread* that continuously listens for new Truss Link connections. The start time of this thread is stored as a UNIX timestamp and is subsequently used as the reference time by all connected links, enabling all devices to use a synchronized clock. Every time a connection from a Truss Link is accepted, a new *Link Handler Thread* begins, and the *Listening Thread* waits for the next Truss Link to connect once more. A lookup table holds all the initiated *Link Handler threads* so that *Controller scripts* can use them to communicate with Truss Links.

The *Link Handler Thread* is responsible for all communications with a link after the initial socket is connected by the *Listening Thread*. It executes the following process:
1) Receive *Hello package* containing Truss Link's device ID
2) Send *Time Epoch package,* which holds the start time UNIX timestamp in it
3) Enter the main update loop:
   a) Receive *Update package* via TCP connection and verify CRC-15 checksum
   b) Update local Truss Link values from package
   c) If socket disconnects or no package is received in 10 seconds, then exit loop
4) Close socket connection cleanly
5) End thread

The above thread only uses the socket corresponding to the Truss Link to receive data. Data sending occurs to all Truss Links from the same socket. For this purpose, the data is encoded using RMLP and then sent via a TCP connection. The Truss Link status will be updated on the server by the Truss Link asynchronously via update packages received by the corresponding *Link Handler Thread*. The message is considered successful if the checksum indicates that the command was received without corruption. Otherwise, the process is repeated.

On the Server, the experiment script can send commands to the Truss Link (instantly) in the main thread while the Link Handler receives update packages asynchronously in an independent thread. Since one thread only reads from the socket, while the other thread only writes to the socket, the implementation is thread-safe, and errors due to parallel usage of the same resources are avoided.

This modular software structure allows the experiment script to include topology controller classes, such as triangle, tetrahedron, or ratchet tetrahedron controllers. These classes allow robots made from Truss Links to be treated as cohesive systems and provide coordinated topology motion functions that assume a specific assembled structure. For example, an experiment script can call a single function from a triangle topology controller class to make a triangle crawl.

An operator used a keyboard interface for our physical experiments to trigger open-loop scripts or send servo commands to the robot. The controller has a keyboard interface that allows the operator to use a topology controller automated movement pattern to make Truss Links crawl, a tetrahedron topple or crawl, or walk a ratchet tetrahedron in discrete directions (see Fig. S4). Individually selected links can be controlled by holding down the number keys corresponding to the Truss Link IDs while simultaneously using the arrow keys to expand or contract the servos of the Truss Link or execute a crawling script in a specific direction.

Experiment environment specifications

The experiment environment was constructed from 25x25 and 25x50 aluminum extrusions. We included a picture of the full setup, including LED lights, in Fig. S1. We covered the environment using a single, continuous strip of carpet.

Truss Link firmware

Truss Links are controlled by a WIFI-enabled Particle Photon microcontroller (Photon). The Photon meets our power and computational requirements and provides a WIFI interface. The Photon allows us to standardize the communication and actuation interface across links. The Truss Links operate by receiving commands via WIFI from our server software. This communication is conducted using our custom protocol. Unlike the server software, the firmware running on the Photon is written in C++, and is computationally lightweight to respect the microcontroller's computational limitations. Fig. S2 illustrates the firmware's process flow.

*Executing Commands:* Due to the nature of the constant loop system, commands are executed at the main loop frequency. As a result, motor commands like sinusoidal motion and position-velocity commands need to be implemented to work in a loop instead of executing until the motion's completion. This was done using *Bang-Bang Motor Control* in combination with *Closed-loop Command Execution.*

*Bang-Bang Motor Control:* A limitation of the Actuonix L-12I motor is that it only physically moves when the difference between actual elongation and desired elongation is greater than 5mm. Thus, when the motor is instructed to move in increments smaller than 5mm, it doesn't respond. This makes it difficult to implement sinusoidal motion patterns and low-velocity movements, as the motor actuates at full speed in a repetitive start/stop manner when presented with gradually increasing position commands. To override this 5mm threshold and force motor movement, we decided to either send a maximum or minimum command to the motor while monitoring the potentiometer feedback until the desired length is achieved. The result is a servo that moves discretely in full-speed increments. This allows us to approximate slower continuous motion via small incremental steps. One artifact of this method is the motor oscillations observed in our physical experiment videos, which result from the Bang-Bang Motor Controller over- and under-shooting the target position.

*Closed-loop Command Execution:* For commands that did not require the servo to actuate at its maximum speed, such as during position-velocity control, instead of directly writing the final stroke length to the motor, we calculated intermediate lengths and gradually sent those as commands to the *Bang-Bang Motor Control*. For each movement, the desired elongation is calculated by passing the time passed since the start of the command into a function that gives the desired stroke-length position for that moment in time, which the motor moves towards using bang-bang control. The combination of bang-bang control and the closed-loop design makes the seemingly parallel receiving and executing commands possible and allows the servo to actuate at a slower speed without stalling for noticeable periods of time.

Communication Protocol

Since the Particle Photon runs C++, and all server-side scripts run Python, a common protocol must be established for socket communication. We developed the *Robot Metabolism Link Protocol* (RMLP), a simple package specification defining communication between Truss Links and the server (see Fig. S3).

The RML Protocol uses packages consisting of C++ structs and datatypes. An RMLP Package consists of 3 parts:

*1) Header:* the header contains two bytes, one holding the package type and the other the package length. This allows the program receiving the package to know how many bytes to read, and how to interpret them based on the package type.

*2) Body:* the body varies based on the type of package. Optimizations were made to minimize the sizes of all types of packages. The body is usually on the order of 5 bytes which keeps data traffic lean even at high frequencies.

*3) Footer:* the footer contains a 16-bit CRC-15 checksum of the package. CRC-15 was picked due to its concise and efficient implementation that can effortlessly run on the Photon. Packages with incorrect checksums can be ignored by the receiver.

The header, body, and footer are appended to create a complete RMLP Package, as illustrated in Table S1.

Manual controller interface

This section explains how the manual control interface for the Truss Links works. The controller employs topology or "species" controllers in the background to control crawling diamonds-with-tail, crawling and toppling tetrahedrons, and ratchet-crawl for ratchet tetrahedrons. These topology controllers assume a specific connection pattern and only work if the structure to be controlled matches the connection pattern assumed by the controller.

The controller script waits for all links to connect before allowing the operator to send commands. The links are sorted based on their Truss Link ID and numbered, this directly corresponds to their position within a diamond-with-tail, tetrahedron, or ratchet tetrahedron. Thanks to this numbering, one can determine in advance which Truss Link will assume which position inside one of those structures, by first sorting the physical links to be used in order of ascending Truss Link ID, and then assembling the shape in the correct order.

Below is a detailed overview of the commands and functionality of the keyboard control interface introduced, shown in Fig. S4:
- 1~9: Select Truss Links by pressing the corresponding number keys on the keyboard. You can press multiple keys simultaneously to select multiple links. Truss Links are mapped to the number keys in ascending order based on their Truss Link IDs. The "1" key is associated with the smallest link ID number, and the "9" key corresponds to the link with the highest Truss Link ID number, assuming nine Truss Links are connected to the server. Pressing 0 will also select the Truss Link with the smallest ID number.
- -/+: Fully contract/expand all Truss Links. When pressing "+", also need to press SHIFT.
- ↑/↓: Expand or contract selected servos on the selected Truss Links.
- ←/→: Select servo 0 or servo 1. If both are pressed, both servos are selected.

- NumLock: Enter *single Truss Links crawling mode*. In this mode, pressing number keys no longer selects the Truss Links to execute commands but directly toggles on/off the crawling for selected Truss Links.
- / *: Set the crawling direction to servo 0 or servo 1 in *single Truss Links crawling mode*.
- s: Clear all sticky commands and stop sending commands to all Truss Links. If a gait is being executed, interrupt the gait.

Preset Gaits
- c, v, b: triangle, tetrahedron, diamond-with-tail crawling
- d, f, g: tetrahedron ratchet crawling
- t, y, u: topple tetrahedron
- o, p: rotate triangle in counterclockwise/clockwise direction
- r: reset ratchet crawling

The below bullet points provide a few examples showing how a desired action corresponds to keys pressed on the controller interface.
- Expand Truss Links 1 fully: press '1', '←', '→', '↑' simultaneously.
- Contract servo 0 of Truss Links 1 and Truss Links 2: press '1', '2', '←', '↓' simultaneously.
- Make Truss Links 1 crawl in the servo 0 direction and make Truss Links 2 start crawling in the servo 1 direction:
    - Press NumLock to enter single Truss Links crawling mode
    - Press '/' to select servo 0 direction or '*' to select servo 1 direction
    - Press '1' to start Truss Links 1 crawling
    - Press '2' to start Truss Links 2 crawling
- Stop tetrahedron crawling and fully contract all Truss Links: press 's' then '-'

Simulation

To study the Truss Links' dynamics and topology formation probabilities, we implemented a PyBullet based simulation environment. Since PyBullet does not provide a native implementation of magnets, we developed our own. We broke down the magnet implementation into two steps: computing pairs of interacting magnets and calculating and applying magnet forces.

Computing magnet interactions is the main bottleneck in our simulation. Evaluating each magnet's interaction with every other magnet is an $O(N^2)$ time complexity problem where N is the number of magnets present in the simulation: mathematically expressed $\binom{N}{2} = \frac{N(N-1)}{2}$. To ensure physical accuracy despite the high magnet forces, magnet interactions must be computed at each simulation timestep i.e., 240 times a second.

Our empirical tests showed that the forces between two magnets were negligible beyond a 14cm distance, even on lower friction experiment surfaces. We exploited this property to develop a quicker algorithm to solve the magnet interaction problem by omitting magnet calculations for magnet pairs further than 14cm apart. We arrived at our matrix-based search-and-apply magnet computation method after exploring various other approaches, including square and triangular occupancy grids to identify interacting magnet sub-groups.

The activation $a_i \in [0, 1]$ of a magnet simulates how the magnet force decreases when the magnet gets retracted inside the connector during the detachment process. An activation of 0 represents a full retraction, and 1 represents no retraction. Hence, for magnets $M_1$ and $M_2$, with respective activations $a_1$ and $a_2$, and separation distance $r$, the force $F$ is described by:

$$|F| = k\frac{a_1 * a_2}{r^2}$$

The problem of computing magnet interactions can be approximated by summing up the resultant force vectors for all interacting pairs of magnets. For example, for a magnet $M_1$ interacting with magnets $M_2$, $M_5$, and $M_7$, the total force is given by

$$\vec{F_{M_1}} = \sum_{i \in \{2,5,7\}} k\frac{a_i * a_1}{\left|\vec{P_i} - \vec{P_1}\right|^3}(\vec{P_i} - \vec{P_1})$$

where $\vec{P_i}$ and $a_i$ represent the position and activation of $M_i$ respectively. In the above formula, $\vec{d_{1,i}} = (\vec{P_i} - \vec{P_1})$ is the direction vector from $M_1$ to $M_i$, and thus its norm is the Euclidian distance between the two magnets.

This approximate method overestimates the force applied to each connector when more than two connectors interact with each other. Thus, the overall attraction to a location increases linearly with the number of connectors at that location. Further, the method doesn't account for magnet polarity, the moment of inertia of the magnet sphere, and the friction between the magnet sphere and the magnet holder. Despite these simplifications, we empirically observed that the simulated Truss Link behavior matched the real-world Truss Link behavior well in simulated experiments with a small number of Truss Links that operated in a high-friction environment.

By exploiting features in Python's NumPy library, distances between all points of magnets can be calculated efficiently. Given a list of $n$ magnets, we construct a $n$x3 matrix $X$ as

$$X = \begin{bmatrix} \vec{P_1} \\ \vec{P_2} \\ \vdots \\ \vec{P_n} \end{bmatrix}$$

Now, we convert the matrix $X$ into a 3-dimensional matrix $X^0$ with dimensions $n$x1x3, which retains the original positions of all elements, but adds a new axis in our matrix. Although redundant at first, this will now let us exploit NumPy's broadcasting functionality, which defines how matrices with different dimensionalities interact during arithmetic operations. We will define a new matrix $D0$ as

$D^0 = X^0 X$

In strict mathematical notation, this is prohibited as the dimensions of $X0$ and $X$ do not match, however, NumPy interprets this as the following sets of operations between *two 3-d matrices of the same dimensions*:

$$D^0 = X^0 - X = \begin{bmatrix} \begin{bmatrix} \vec{P_1} \\ \vdots \\ \vec{P_1} \end{bmatrix} \\ \vdots \\ \begin{bmatrix} \vec{P_n} \\ \vdots \\ \vec{P_n} \end{bmatrix} \end{bmatrix} - \begin{bmatrix} X \\ \vdots \\ X \end{bmatrix} = \begin{bmatrix} \begin{bmatrix} \vec{P_1} \\ \vdots \\ \vec{P_1} \end{bmatrix} \\ \vdots \\ \begin{bmatrix} \vec{P_n} \\ \vdots \\ \vec{P_n} \end{bmatrix} \end{bmatrix} - \begin{bmatrix} \begin{bmatrix} \vec{P_1} \\ \vdots \\ \vec{P_n} \end{bmatrix} \\ \vdots \\ \begin{bmatrix} \vec{P_1} \\ \vdots \\ \vec{P_n} \end{bmatrix} \end{bmatrix}$$

We subsequently square each element of the matrix $D0$ and sum all components of each vector to get a n × n matrix $N_2^0$ where the $N_2^0[i,j]$ is the square of the norm between magnets $M_i$ and $M_j$. Then we apply NumPy function *argwhere* on $N_2^0$ where the squared norm is smaller than the square of our maximum interaction distance to get a $l$x2 matrix containing pairs of indices whose magnet interaction is significant. We then remove entries in that matrix where the *i* index is less than the *j* index to ensure that the indices of each pair of proximate magnets is recorded only once using the *arghwere* function once more. We now have all the lists of pairs of magnets that are close enough to interact and the squared norms of those pairs in $N_2^0$.

We use the Fast Inverse Square Root algorithm which uses advanced bit manipulation and Newton Iteration to speed up the approximation of the inverse square root (see *lomont.org/papers/2003/InvSqrt.pdf*). Using the fast inverse square root algorithm on every element of $N_2^0$ simultaneously, we calculate the inverse of the norms of the rows of $D^0$ very efficiently. We can subsequently apply the formula for the resultant force simultaneously for all pairs using NumPy operations. This algorithm exploits NumPy parallelization and bit manipulation, outperforming all other previous methods we tested on our hardware: a workstation with an AMD Threadripper 2950X CPU, 64GB of RAM, and three NVIDIA RTX 2080Ti GPUs. The speed improvement of this matrix-based magnet computation method over other methods we tried becomes more significant as the number of magnets in the simulation increases.

Collision computations are another bottleneck in simulation, especially if the collision bodies used are complex. To address this issue and speed up the simulation, we have found shape primitive terrains to be a happy medium between accuracy and performance. For example, arranging boxes in a circle to form a circular playpen is easy and avoids using a complex body mesh. The primitives supported in PyBullet are box, sphere, cylinder, capsule, plane, and multi-sphere. We designed the experiment environment shown in Fig. 3 using only boxes and a cylinder in simulation. The benefit of using primitives instead of a heightfield for this application is that the collision computation between flat surfaces is less complex than between the fine triangulated mesh of a heightfield. We found primitive terrains to run significantly faster than their heigh-field-based counterparts. Furthermore, heightfields only approximate perpendicularly intersecting faces and, thus, may represent such shapes with less accuracy than an arrangement of boxes, especially if the heightfield resolution is reduced to increase performance.

Morphological representation and tracking in simulation

To study the morphological development of Truss Links in simulation, we needed an efficient approach to represent the topological and geometrical configuration while also encoding the connection information between them. An efficient representation is crucial to identifying and tracking newly formed structures and analyzing different morphologies. A Morphology refers to the physical arrangement of Truss Links wherein each Truss Link is defined by its pose (i.e., position, orientation) and 2 servo positions. We implemented a "Structure Graph" representation to encode a physical Truss Link morphology as a digital object to represent this physical configuration.

Since the Truss Link is a rigid body, we can determine its exact pose from the positions of the two magnets embedded in its connectors and servo positions. To identify how Truss Links are connected, we only need the positions of its magnets. We know empirically that Truss Link connector tips within two inches of each other snap together, and we can use this information to identify connections between magnets.

Given a list of *n* magnets, we can compute a list of all pairs of magnets within a certain threshold distance from each other. This can be done in *O(n)* time using an occupancy grid to identify nearby magnets. Let us call this new list *connections,* assuming that magnets within the threshold distance are necessarily physically connected. In addition, we append all self-connections i.e., ($m_i$, $m_j$) $\forall$ i $\in$ {1, $\cdots$, $n$} to the *connections* list. Now, *connections* can be treated as a list of edges in an undirected graph $G_m$ describing physical magnet contacts. We then identify all independent subgraphs in $G_m$ by repeatedly performing a breadth-first search. Each independent subgraph in $G_m$ represents a subset of connected magnets. Based on this physical interpretation, let us consider each independent subgraph in $G_m$ a geometric vertex of our robot morphology. We define the vertex position as the mean position of the magnets it comprises. We construct a new graph called the *structure graph* using these newly found vertices. The *structure graph* represents the robot's morphology. The nodes of the *structure graph* are the vertices we identified. We add an edge for each link connecting two vertices. As a result, we get a graph where the nodes represent clusters of connected magnets, and the edges represent the Links themselves, thereby representing the structure created by interconnecting Links.

The graph representation is ideal because it allows us to identify structures with identical morphological properties regardless of where the morphology is located, which links it comprises, and how the links are oriented. Consider two tetrahedron morphologies comprising 12 links. Now, we can compute the *structure graph* for each tetrahedron and algorithmically verify that both structures are tetrahedrons by checking if an isomorphism exists between them. Identifying isomorphism can be done regardless of which unique links in which orientations comprise the robot and in which directions these links are connected. The Truss Links orientation is relevant to the controller since motor commands need to be sent to the correct motor. Once an isomorphism is identified, we can algorithmically find which directed Truss Link in the first tetrahedron is equivalent or (isomorphic) to which directed Truss Link in the second tetrahedron. If an isomorphism exists, we could use this information to control both morphologies identically by applying the same controller.

Another benefit of the graph representation is that two isomorphic graphs hash to the same value. We used the Weisfeiler-Lehman Hash (WL-hash) for this purpose. The ability to hash graphs enables us to *name* morphologies as we encounter them automatically. *Naming* refers to the action of giving a unique string representation to a particular structure (for example, a triangle, a tetrahedron, or a sea-urchin-like structure formed from 20 Truss Links). In future research, we plan to automatically generate *Phylogenetic Trees* or *Finite State Transition Models* based on the sequence of hashes through which morphologies tend to develop and use them to associate learned controllers for each morphology.

One shortcoming of the WL-hash is that it causes hashing collisions for graphs comprising independent subgraphs. For example, a connected cycle comprising six Truss Links would produce the same hash as two separate triangles hashed as one graph. Thus, to accurately hash all structures formed in a simulation, one has to iterate through all the connected components of the graph and compute their WL-hash separately.

**Figure S1: Photograph of the physical experiment environment.** The experiment environment was built from aluminum extrusions and occupies an area of 4.3 meters by 1 meter.

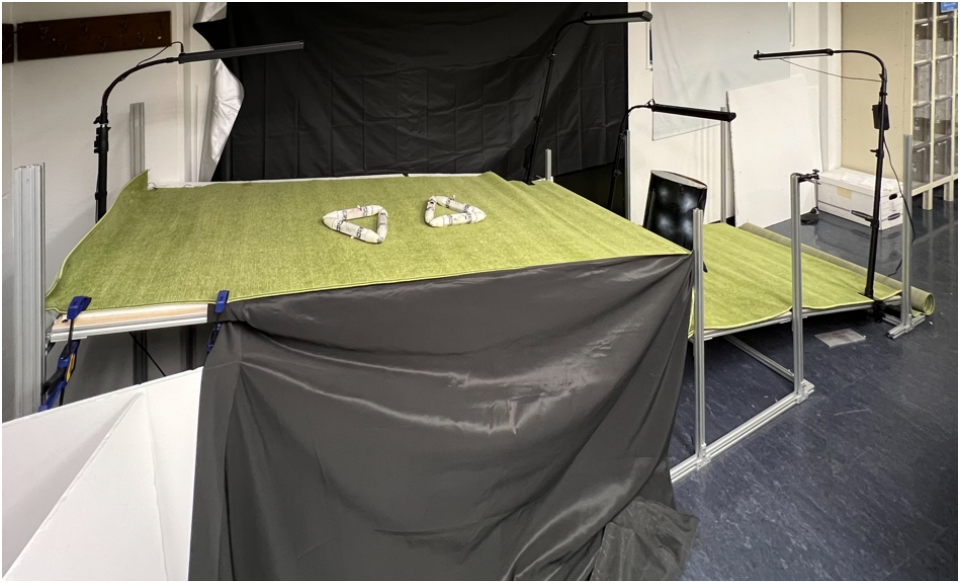

**Figure S2: Flow diagram showing the states and transitions of the Truss Link's firmware.**

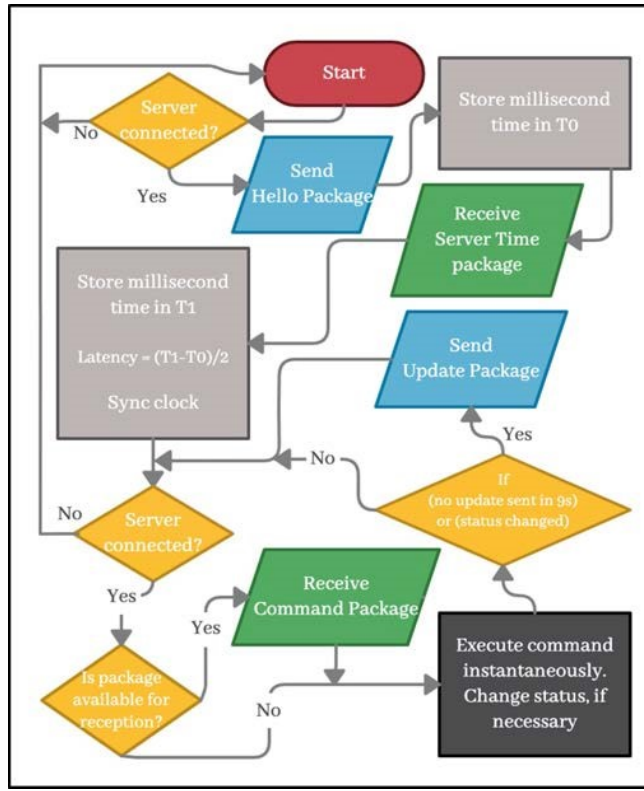

**Figure S3: Truss Link system communication diagram.** The diagram shows how most computational loads are processed offboard on a laptop computer, with the necessary commands being shared with all Particle Photon microcontrollers of the Truss Links via the Robot Metabolism Link Protocol (RMLP) over an active WIFI connection.

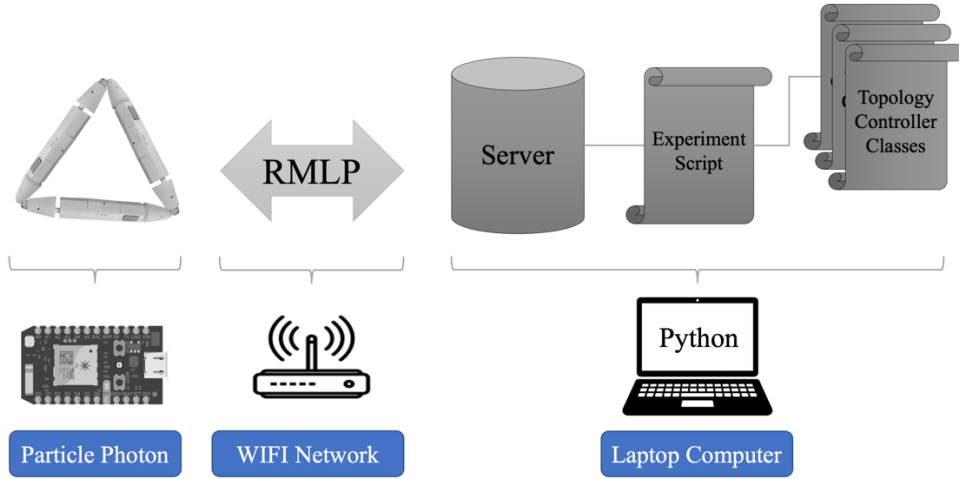

**Figure S4: Overview of the key mappings used in the manual control interface.**

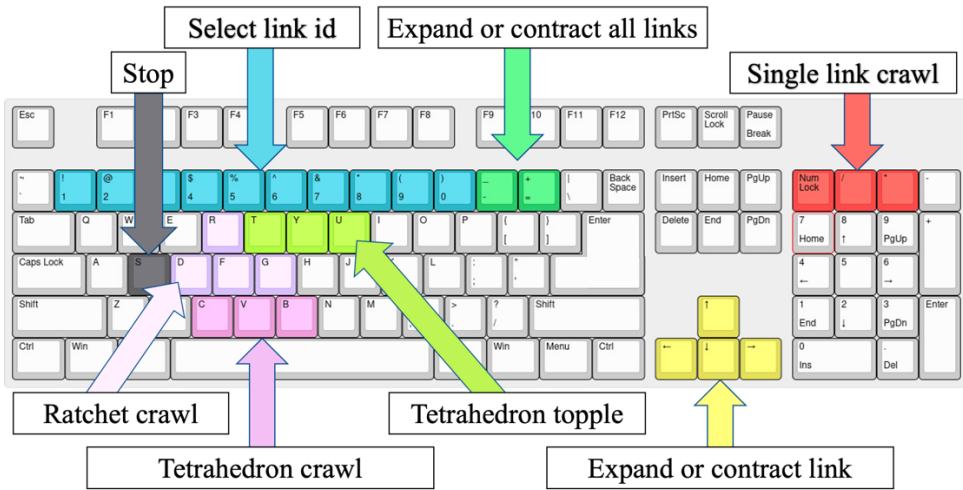

**Table S1:** Each RM Protocol package has a header, a body, and a footer. The header is 2 bytes in size, and contains the data size, and type. The body ranges in size based on the message type, and the footer is 2 bytes. The header specifies the size of the body.

|          | Header    |          | Body           | Footer   |
|----------|-----------|----------|----------------|----------|
| Size     | 1 byte    | 1 byte   | ~5 < 256 bytes | 2 bytes  |
| Contents | data size | Pkg Type | Data           | Checksum |

**Movie S1.**

Four developmental transformations

This video shows the four developmental transformations from seven independent links to a ratchet tetrahedron topology. At each stage, the robot grows by absorbing more material. First, six independent links combined into a 3-pointed star and a triangle. The triangle then integrates the three-pointed star and forms a diamond-with-tail configuration. Second, the diamond-with-tail crawls over the ledge in our experiment environment and folds itself into a tetrahedron shape. Finally, the tetrahedron picks up a single found Truss Link, and uses it like a walking stick in a ratchet tetrahedron configuration. It then goes on to propel itself down the slope of the ramp in large sliding strides. This video demonstrates the Robot Metabolism's ability to grow within its lifetime and change both its topology and its functionality.

**Movie S2.**

Robot consumes material and self-improves at every stage

This video shows how each newly formed topology gained an ability unavailable to the previous one. First, a single Truss Link and a triangle try to circumvent an obstacle. The single Truss Link gets stuck since it lacks the ability to steer, while the triangle crawls around the obstacle with ease. Second, the triangle and the diamond-with-tail attempt to overcome a 2.5cm threshold with a bank on either side. Due to its long body, the diamond-with-tail can push itself over the threshold, while the triangle cannot shift its weight enough to overcome this obstacle. Third, a diamond-with-tail and a tetrahedron attempt to overcome a 2.5cm square aluminum extrusion. The diamond-with-tail cannot lift its tip and thus fails to overcome the obstacle, while the tetrahedron topples itself onto the aluminum extrusion and then crawls down on the other side. Finally, a tetrahedron and a ratchet tetrahedron face off in a race on a sloped surface. The ratchet tetrahedron arrives at the end of the slope while the tetrahedron just crosses the halfway line. This video demonstrates how growing and adapting allows even simple robots to gain a significant advantage.

**Movie S3.**

Random diamond-with-tail formation in simulation

This video shows a rendering of a randomly moving set of six Truss Links forming a diamond-with-tail. We added walls to the experiment environment to ensure the links don't fall outside the experiment environment. The Truss Links in this video execute random motions based on a Fourier series initialized with random parameters sampled from a functional range of parameters. Due to the greedy attachment style of the Truss Links, random motion is likely to lead to the formation of new connections if links are nearby. The video rendering is sped up. In reality, Truss Link servos take 15-20 seconds to expand, varying slightly based on the battery voltage provided.

**Movie S4.**

Replace "dead" Truss Link and recover shape after impact

Truss Links are programmed to fully contract and detach when their battery voltage drops below a critical level. This video shows a ratchet tetrahedron robot that finds a new Truss Link and shortly thereafter loses its ratchet Truss Link as it would if the link "died," i.e., ran out of battery. The ratchet tetrahedron robot then uses the newly found link as its ratchet link triangle. Next, a three-pointed star, a triangle, and a diamond-with-tail robot crawl off a drop and recover their previous topology.

**Movie S5.**

Assisted tetrahedron formation

This video shows a ratchet tetrahedron on a platform, placing itself above a hole in the platform, while a three-pointed star and a triangle connect at a single point. This newly formed structure crawls down the slope while the dangling Truss Link of the ratchet tetrahedron fishes for the central vertex of the planar structure below. Once connected, the ratchet tetrahedron uses the ratchet link's body to support itself on the hole's edge and then elevates the links below by contracting one of its servos. You'll notice the servo oscillating in and out since the operator is trying to avoid entirely contracting the link since that would detach the lifted payload. Simultaneously, the two free links of the three-pointed star are shuffling their way down the slope to connect to the two free vertices of the triangle. Once they connect and the tetrahedron is formed, the above ratchet Truss Link fully contracts its one servo to let go of the tetrahedron. Both the tetrahedron and the ratchet tetrahedron crawl away.

**Raw experiment data and analysis scripts**

The *supplementary_data.zip* file contains two sub-directories: *MultiTopologySpeedAnalsyis* and *RandomTopologyFormation*. The first contains the data and the Google Colab notebooks used to produce the results presented in Fig. 4 and Table 1. The second contains the simulation results and the analysis Jupyter Notebook used to calculate the formation probabilities in Fig. 5.